\documentclass[accepted, specialissue]{melba}

\usepackage{amsmath,amsfonts}

\usepackage{comment}
\usepackage{bm}
\usepackage{arydshln}
\usepackage{xcolor}
\newcommand{\jan}{}

\melbaid{2026:008}  
\doi{10.59275/j.melba.2026-1ba1}
\melbaauthors{Boysen et al.}  
\email{jan.boysen@dfki.de}
\volume{2026}
\firstpageno{148}  
\melbayear{2026}  
\datesubmitted{2025-07}  
\datepublished{2026-03}  

\melbaspecialissue{MELBA–BVM 2025 Special Issue}
\melbaspecialissueeditors{Andreas Maier, Thomas Deserno, Heinz Handels, Klaus Maier-Hein, Christoph Palm, Thomas Tolxdorff, Katharina Breininger}

\ShortHeadings{Biophysics-Enhanced Respiratory Motion Modeling}{Boysen, Uzunova, Handels, Ehrhardt}

\title{Biophysics-Enhanced Neural Representations for \\ Patient-Specific Respiratory Motion Modeling}

\author{
	\firstname Jan \surname Boysen\aff{1}\orcid{0009-0005-1308-6696},
	\firstname Hristina \surname Uzunova\aff{3}\orcid{0000-0001-8854-3517},
	\firstname Heinz \surname Handels\aff{1,2}\orcid{0000-0002-3499-4328},
	\firstname Jan \surname Ehrhardt\aff{1,2}\orcid{0009-0004-6804-5587}
}

\affiliations{
	\num 1 \addr German Research Center for Artificial Intelligence (DFKI), Lübeck, DE \\
	\num 2 \addr Institute of Medical Informatics, University of Lübeck, Lübeck, DE \\
    \num 3 \addr Clinic for Orthopedics and Orthopedic Surgery, University Medicine Greifswald, Greifswald, DE
}

\abstract{
A precise spatial delivery of the radiation dose is crucial for the treatment success in radiotherapy. In the lung and upper abdominal region, respiratory motion introduces significant treatment uncertainties, requiring special motion management techniques. To address this, respiratory motion models are commonly used to infer the patient-specific respiratory motion and target the dose more efficiently. In this work, we investigate the possibility of using implicit neural representations (INR) for surrogate-based motion modeling. Therefore, we propose physics-regularized implicit surrogate-based modeling for respiratory motion (PRISM-RM). Our new integrated respiratory motion model is free of a fixed reference breathing state. Unlike conventional pairwise registration techniques, our approach provides a trajectory-aware spatio-temporally continuous and diffeomorphic motion representation, improving generalization to extrapolation scenarios. We introduce biophysical constraints, ensuring physiologically plausible motion estimation across time beyond the training data. Our results show that our trajectory-aware approach performs on par in interpolation and improves the extrapolation ability compared to our initially proposed INR-based approach. Compared to sequential registration-based approaches both our approaches perform equally well in interpolation, but underperform in extrapolation scenarios. However, the methodical features of INRs make them particularly effective for respiratory motion modeling, and with their performance steadily improving, they demonstrate strong potential for advancing this field.}


\begin{document}

\twocolumn[\maketitle]

\section{Introduction}
\enluminure{T}{he} aim of radiotherapy is to accurately deliver a high dose of radiation to the tumor while sparing as much healthy tissue as possible. In tumor treatment in the lung and upper abdomen, respiratory motion is one of the main sources of error, and respiratory motion management such as respiratory gating or tracking is required to compensate for tumor motion during treatment~\citep{dhont_image-guided_2020}.
\jan{The ideal scenario for managing respiratory motion is to directly observe and track the tumor and organs at risk (OARs) during radiation therapy. 
However, achieving this requires suitable imaging technologies integrated into the linear accelerator that either provide functional information, such as PET~\citep{fiorina_detection_2021}, or high soft tissue contrast, such as MRI~\citep{cervino_mri-guided_2011}. 
The clinical application of techniques like MRI-guided radiation therapy (MRIgRT) is currently limited by the need for additional infrastructure, increased treatment and personnel costs, and the lack of medium- to long-term clinical outcome data~\citep{keall_integrated_2022}. 
Less complex and more cost-effective linac-integrated imaging techniques such as MV/kV imaging enable sufficiently reliable and robust marker-less tracking only for specific cases and tumor locations~\citep{zhang_design_2018}. 
An alternative to direct imaging and tracking of tumor and OARs is indirect monitoring using internal or external surrogate signals. 
Examples of internal surrogate signals include implanted fiducials tracked by kV imaging or electromagnetic transponders~\citep{cho_first_2009, bertholet_real-time_2019}, while external surrogate signals include spirometers, pressure belts, or optical tracking systems. 
In this work, we rely on external surrogate signals, although the proposed methodology is equally applicable to internal surrogate signals.}

\jan{Then, based on a surrogate signal that can be acquired during treatment, deformable motion models allow for the estimation of respiratory motion in an entire anatomical region, including the tumor and nearby OAR}~\citep{mcclelland_respiratory_2013}.
Such models are usually generated based on previously acquired 4D image data using a sequential process including two-stages: first, the motion is estimated from the imaging data using deformable image registration and second, a correspondence model approximates the relationship between the surrogate signal and the motion using regression modeling~\citep{mcclelland_respiratory_2013,wilms_multivariate_2014}.
Later work~\citep{mcclelland_generalized_2017,huang_resolving_2024} proposed an integrated model that jointly optimizes image registration and correspondence model fitting, improving the reconstruction of 4D image data. The majority of approaches  assume simple linear or polynomial relationships between the surrogate signals and local deformations.

The estimation of motion from 4D image data is usually done by pairwise registration of the dynamic 3D image frames to a fixed image, acquired at a reference breathing state~\citep{ehrhardt_4d_2013}. A variety of optimiza-tion- and learning-based registration approaches has been applied specifically to thorax images, see~\cite{xiao_deep_2023} for an overview. Recently, implicit neural representations (INRs) have gained attention in the field of medical image registration~\citep{sun_medical_2024}, performing on par with the state-of-the-art for deformable lung registration~\citep{wolterink_implicit_2022,zimmer_towards_2024}. INRs offer a continuous representation of deformation, and their capacity for analytical differentiation provides significant benefits for implementing PDE-based regularizers and loss functions~\citep{chen_implicit_2023}.
%
Closely related to physics-informed neural networks (PINNs)~\citep{raissi_physics-informed_2019}, biophysical constraints can be introduced into the registration model as proposed by~\cite{arratia_lopez_warppinn_2023} and, moreover, INRs naturally fit into a diffeomorphic registration setting~\citep{sun_topology-preserving_2022,tian_nephi_2025,han_diffeomorphic_2022,zhang_inr-lddmm_2023,sun_medical_2024}.
Another line of research uses INRs for motion-compensated image reconstruction and benefits in particular from the implicit continuity regularization of INRs and the ability to reconstruct from sparsely and irregularly sampled data~\citep{feng_spatiotemporal_2025,shen_nerp_2024,zhang_dynamic_2023}.
Although some of these approaches include the cardiac or respiratory phase as a temporal dimension, the motion is only implicitly modeled. 
Despite the success of the approaches in the context of image registration and motion-compensated image reconstruction, there is rarely any work utilizing INRs for surrogate-based motion modeling.

In this work, we investigate implicit neural representations for creating patient-specific motion models that represent the lung deformation
over an entire breathing cycle. Specifically, based on a 4D image dataset with associated surrogate signals, we aim to learn a model that is able to estimate the internal motion state from unseen surrogate signals. 
This work builds upon our previous conference paper~\citep{palm_surrogate-based_2025}, in which we developed an integrated INR-based motion model that uses spatial position and surrogate signals as input dimensions to predict local motion. The network is capable of learning complex functions to model the relationship between surrogate signals and internal breathing motion. It learns the respiration-induced motion deformations from sparsely sampled image data by registering images from different \jan{breathing states to a reference state}. 

This paper extends our previous work in several aspects: First, a much more detailed mathematical derivation and description of the approach is presented  in the context of surrogate based motion modeling. More importantly, based on the transport equation, we extend our INR-based approach to a diffeomorphic, time-continuous respiratory motion model that no longer requires a fixed reference image for pair-wise registration. Instead, we introduce a group-wise diffeomorphic INR-based registration, enabling motion estimation between arbitrary respiratory phases. To determine the motion from one breathing phase to another, our model traces the trajectories along the breathing curve. Model training is, therefore, performed by sampling along the respiratory trajectories between arbitrary image pairs. 
Besides the physics-based spatial regularization introduced in~\cite{palm_surrogate-based_2025}, additional temporal constraints are employed to ensure smooth and plausible motion trajectories. 
The motivation for these substantial adaptations of our approach is twofold: First, we aim to improve the robustness by increasing the number of available image pairs for training from $n-1$ when using a fixed reference image to $n(n-1)$ with arbitrary image pairs. Second, we address the extrapolation limitations commonly observed in implicit neural representations and also reported in~\cite{palm_surrogate-based_2025}. 
Since surrogate signals are recorded continuously while images are only available at a discrete amount of time points, our trajectory-based sampling scheme together with the temporal smoothness constraint forces our model to learn plausible motion information, even for surrogate signals lying outside of acquired image information. 

We call our adapted and improved approach a physics-regularized implicit surrogate-based modeling for respiratory motion (PRISM-RM), and provide evaluation experiments on 3D+t (4D) CT images of lung cancer patients with corresponding surrogate signals to show the advantages of these adaptations, especially in extrapolation scenarios. Further, we investigate the introduced spatial and temporal regularizations in an ablation study.


\section{Methods}
For surrogate-based modeling we consider the following scenario: given is a 4D image dataset $\mathcal{I} = \{I_j\}_{j=0}^{n-1}$ consisting of $n$ 3D images $I_j: \Omega\rightarrow\mathbb{R}\ (\Omega\subset\mathbb{R}^3)$ acquired at time-points $t_j\in[0,T]$ and a time-continuous surrogate signal  $\mathbf{s}:[0,T]\rightarrow\mathcal{S}\ (\mathcal{S}\subset\mathbb{R}^{n_{s}})$ of dimension $n_{s}$, where $n_s$ is the surrogate signal dimensionality, such that $\mathbf{s}(t_j)$ corresponds to $I_j$.

\begin{figure*}[t]
    \centering
    \includegraphics[width=\textwidth]{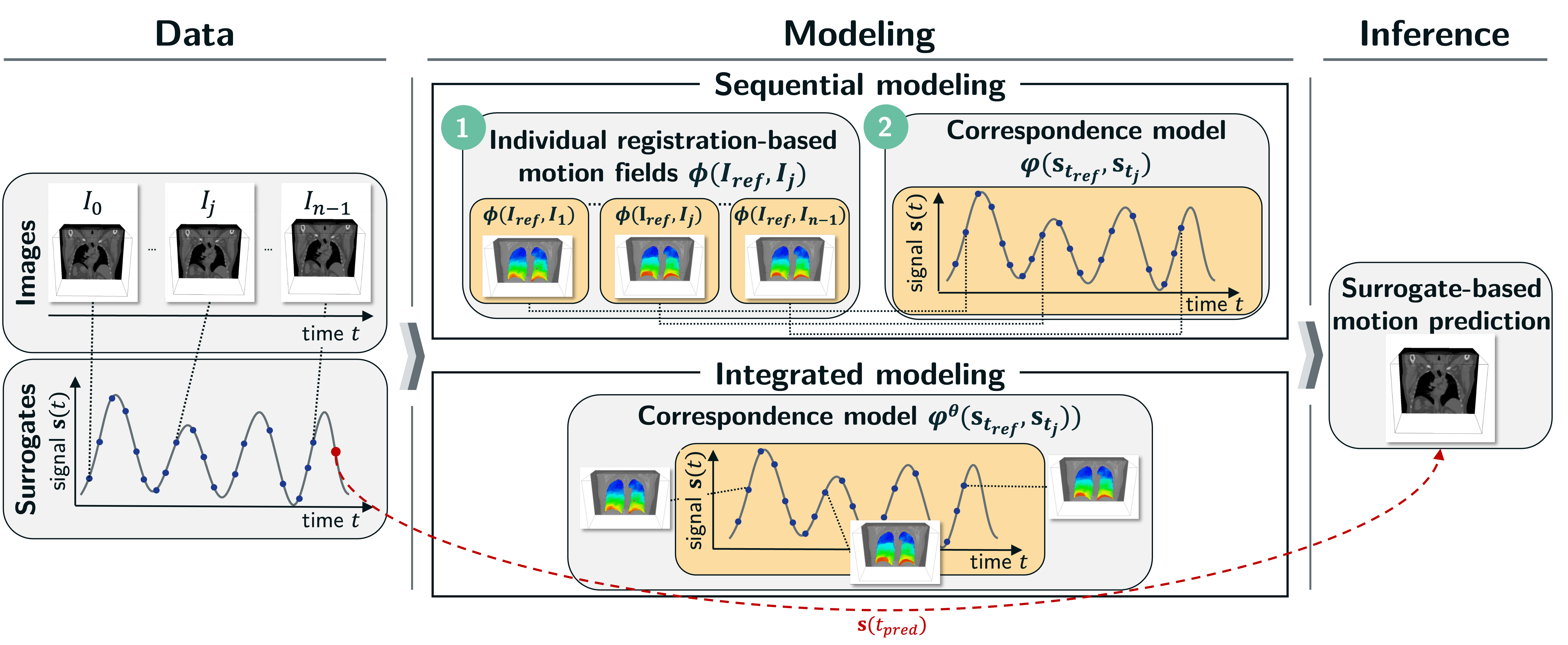}
    \caption{Sequential and integrated respiratory motion modeling using image data with corresponding surrogate signals over time.}
    \label{fig:integrated_approach}
    \label{fig:twostage_model}
\end{figure*}

\subsection{Background on Surrogate-Based Motion Modeling}
Surrogate-based motion models aim to predict internal organ motion based on externally measured surrogate signals $\mathbf{s}$. More formally, the function to be determined is
\begin{equation}\label{eq:surrogate}
    \varphi:\Omega\times\mathcal{S}\rightarrow\Omega \quad \mathbf{\hat{x}}=\varphi^{\bm{\theta}}(\mathbf{x},\mathbf{s}).
\end{equation}
The function $\varphi^{\bm{\theta}}$ is parametrized by $\bm{\theta}$
and predicts the position of a particle $\mathbf{\hat{x}}$ initially positioned at $\mathbf{x}\in\Omega$ given the  measured signal $\mathbf{s}\in\mathcal{S}$. 
The parameters $\bm{\theta}$ are the coefficients of a regression model that relates the surrogate signals to the internal spatial motion. They are learned from 4D image data acquired together with surrogate signals. 
Usually, patient-specific images acquired before radiotherapy are used to build a correspondence model; however, population-based models can also be created using multiple 4D image datasets~\citep{mcclelland_respiratory_2013,wilms_multivariate_2014,wilms_subpopulation-based_2017}.

\subsubsection{Sequential Respiratory Motion Modeling}\label{subsec:twostagemodel}
Sequential respiratory motion models describe a group of motion modeling approaches that feature two sequential stages~\citep{mcclelland_respiratory_2013}.
First, the respiratory motion is estimated by image registration from the images $I_j$. Second, a correspondence model is fitted mapping the motion with the corresponding surrogate signals $\mathbf{s}(t_j)$ (see Fig.~\ref{fig:twostage_model} for a schematic overview).

In the first stage, all images $I_j$ are registered to a fixed reference state image, e.g. $I_{ref} = I_{0}$, to obtain the transformations $\phi_j:\Omega\rightarrow\Omega, j=1,\ldots,n-1$. In the case of pair-wise deformable image registration, the following objective is optimized:  
\begin{equation}\label{eq:registration_loss}
    \mathcal{L}_j(\phi_j) = \mathcal{D}\left(I_{ref},I_j\circ\phi_{j}\right) + \alpha\mathcal{R}\left(\phi_{j}\right),
\end{equation}
where $\mathcal{D}$ denotes a similarity measure and $\mathcal{R}$ is a regularization term weighted by $\alpha$.

In the second stage, the relationship between the acquired surrogate signals $\bm{s}(t_j)$ and the $n-1$ estimated deformation fields $\phi_{j}$ is learned by fitting a regression model to approximate the deformations, i.e. $\phi_j\approx\varphi^{\bm{\theta}}(\mathbf{s}(t_j))$. 
Most approaches use simple linear or polynomial models combined with standard fitting techniques such as ordinary least squares~\citep{mcclelland_respiratory_2013}. Following the example of \cite{wilms_multivariate_2014}, the deformations are represented by the group exponential map of a stationary velocity field $\phi_j(\mathbf{x})=\exp(v_j)(\mathbf{x})$~\citep{hutchison_log-euclidean_2006}. 
Let $\mathbf{V_j}\in\mathbb{R}^{3m}$ be the matrix of vectorized representations of the velocity fields $v_j$, where $m$ denotes the number of voxels. The parameters $\bm{{\theta}}$ of a linear correspondence model with regression coefficients $\mathbf{B}\in\mathbb{R}^{3m\times n_s}$ and intercept $\mathbf{b}_0\in\mathbb{R}^{3m}$ are given by:
\begin{equation}\label{eq:OLS}
    \bm{{\theta}}=\{\mathbf{{B}}, \mathbf{{b}}_0\} = \arg\min_{\{\mathbf{B}, \mathbf{b}_0\}} \sum_{j=1}^{n-1} \|\mathbf{V}_j - \mathbf{B}\mathbf{s}(t_j) - \mathbf{b}_0 \|_2^2.
\end{equation}
During inference, a newly measured surrogate signal $\mathbf{s'}$ is applied to predict velocities $\mathbf{V}=\mathbf{B}\mathbf{s'} + \mathbf{b}_0$ followed by exponentiation to compute the deformation.
To model the variations in motion between inspiration and expiration (hysteresis), many approaches use separate models for inhalation and exhalation due to the limited expressivity of correspondence models.

\subsubsection{Integrated Respiratory Motion Modeling}\label{subsec:integratedmodel}

In the integrated respiratory motion modeling approach, the idea is to 
directly estimate the parameters $\bm{\theta}$ of the correspondence model from the images $I_j$ and corresponding surrogate signals $\mathbf{s}(t_j)$.
Here, an optimization over the parameters $\bm{\theta}$ of the correspondence model simultaneously adapts the reference image $I_{ref}=I_0$ to all remaining images $I_j, j=1,\ldots,n-1$.
The objective can be defined analogously to Eq.~(\ref{eq:registration_loss}):
\begin{equation}\label{eq:integrated_loss}
    \mathcal{L}(\bm{\theta}) = \sum_{j=1}^{n-1} \mathcal{D}\left(I_{ref},I_j\circ\varphi^{\bm{\theta}}\left(\mathbf{s}(t_j)\right)\right) + \alpha\mathcal{R}\left(\varphi^{\bm{\theta}}\left(\mathbf{s}(t_j)\right)\right).
\end{equation}

In contrast to the sequential approach~(Sec.~\ref{subsec:twostagemodel}) the parameters $\bm{\theta}$ are optimized directly by a gradient-based algorithm, and the registration-based estimation of intermediate deformations $\phi_j$ is avoided; thus, the optimization is carried out simultaneously over all images and surrogates. The integrated modeling approach is shown in Fig.~\ref{fig:integrated_approach}.

One integrated respiratory motion model was introduced by \cite{mcclelland_generalized_2017}. The authors define a linear correspondence model in order to map the surrogate signals to the parameters of the deformable transformation, such as velocities, displacements or B-spline functions.
Theoretically, the correspondence model can be defined in any differentiable fashion, e.g. with polynomials or spline-based regression, however, this approach requires a strict definition of the relation between surrogate signal and motion parameters.
Omitting the frame-wise estimation of deformations increases robustness and has been applied successfully to, e.g., 4D image reconstruction when only partial data was acquired per time point~\citep{mcclelland_generalized_2017, huang_resolving_2024}.

In our previously published conference paper~\citep{palm_surrogate-based_2025}, we aimed to lift the restrictions of an explicitly defined correspondence model and explore the capabilities of the recently emerged INRs~\citep{sitzmann_implicit_2020}. INRs map spatial coordinates $\mathbf{x}$ (and surrogate values) to their corresponding image intensity by training a dense neural network as a mapping function $f(\mathbf{x})$, enabling continuous spatio-temporal representations, making them invariant to image resolution and size. Especially, for patient-specific applications, INRs have shown great advantages in terms of memory consumption and speed and, thus, pose a promising alternative to conventional deep-learning approaches.

Inspired by \cite{wolterink_implicit_2022}, we propose to use surrogate-conditioned INRs to implicitly represent the displacement-based warping function between a globally fixed reference image $I_{ref}$ and some image $I_j$ with the corresponding surrogate signal $\mathbf{s}(t_j)$:
\begin{equation}
    \varphi^{\bm{\theta}}(\bm{x}, \mathbf{s}) = \bm{x} +f^{\bm{\theta}}(\bm{x, \mathbf{s}}),
\end{equation}
with $\bm{\theta}$ being the parameters of the INR. Notably, this approach still requires a fixed reference image $I_{ref}$ and also to learn separate models for the inhalation and exhalation phases. Additionally, to address the overfitting risk of deep neural networks, we include a physics-inspired spatial regularization inspired by PINNs~\citep{raissi_physics-informed_2019} (Sec.~\ref{subsec:regularizer}).

\subsection{Trajectory-Aware Respiratory Motion Modeling}\label{subec:diffeo_INR}

Following the idea of integrated respiratory motion models (Sec.~\ref{subsec:integratedmodel}), we propose to extend the INR-based integrated respiratory modeling approach by re-formulating it as a reference-free diffeomorphic model.
We assume that for a particle at position $\mathbf{x}\in\Omega$ at time $t_0=0$ and surrogate signal $\mathbf{s}(t_0=0)$, the function $\varphi(\mathbf{x}, \mathbf{s}(t_j)): \Omega\times\mathcal{S}\rightarrow\Omega$ defines the position of that particle at time $t_j$ for a given surrogate signal $\mathbf{s}(t_j)$. More generally and for ease of notation regarding the particle's position, let $\phi_{r,t}(\mathbf{x})$ be the position of a particle at time $t$ with starting position $\mathbf{x}$ at time $r$, in our example $\varphi(\mathbf{x}, \mathbf{s}(t_j))=\phi_{t_0,t_j}(\mathbf{x})$.
To ensure diffeomorphism, the motion $\phi_{r,t}$ is generated as a flow of the time-dependent velocity field $v$ defined by the transport equation:
\begin{equation}\
    \frac{\partial \phi_{r,\cdot}(\mathbf{x})}{\partial t}=v^{\bm{\theta}}(\phi_{r,t}(\mathbf{x}),\mathbf{s}(t)) \quad\text{s.t. } \phi_{r,r}(\mathbf{x})=\mathbf{x}. 
    \label{eq:diffeo}
\end{equation}
The time-dependent velocity field $v^{\bm{\theta}}$ is represented by an INR parametrized by network weights $\bm{\theta}$, and using the spatial coordinate and surrogate as inputs.
To now estimate the motion between two time frames $t_j$ and $t_k$, starting from the given velocity field, an integration over time is required:
\begin{equation}
    \phi^{\bm{\theta}}_{t_j,t_k}(\mathbf{x})=\mathbf{x} + \int_{t_j}^{t_k} v^{\bm{\theta}}(\phi_{t_j,\tau}(\mathbf{x}), \mathbf{s}(\tau))d\tau.
    \label{eq:integration}
\end{equation}
The resulting deformations $\phi^{\bm{\theta}}_{t_j,t_k}$ 
are diffeomorphic at all times $t_j,t_k\in [0, T]$~\citep{younes_shapes_2019}.
Here, we propose to realize integration using an Euler-based scheme with variable step-size and a fixed number of evaluations. Note that this formulation diverges from the widely used scaling-and-squaring-based approaches~\citep{hutchison_log-euclidean_2006,dalca_unsupervised_2019} in two ways: we employ variable integration intervals rather than $[0,1]$, and we employ time-dependent velocity fields, similarly to LDDMM~\citep{beg_computing_2005}.

The time-continuous formulation in Eqs.~(\ref{eq:diffeo})~and~(\ref{eq:integration}) now allows us to reformulate the integrated motion model in Eq.~(\ref{eq:integrated_loss}) to:
\begin{equation}
    \mathcal{L}(\bm{\theta})=\sum_{\substack{(j,k)\in[1,\ldots n]^2\\ j\neq k}} \mathcal{D}(I_j,I_k\circ\phi^{\bm{\theta}}_{t_j,t_k}) + \alpha \mathcal{R}(\phi^{\bm{\theta}}_{t_j,t_k}, v^{\bm{\theta}}),
    \label{eq:optimization_criterion}
\end{equation}
with notation analogous to Eq.~(\ref{eq:registration_loss}).
This reformulated approach has a number of advantages. First, by omitting a fixed reference frame, we can use $n(n-1)$ registration pairs to learn the network parameters instead of only $n-1$ as in the approach in Sec.~\ref{subsec:integratedmodel}. Second, for the calculation of the motion deformations $\phi_{t_j,t_k}$ using the integral in Eq.~(\ref{eq:integration}) the recorded surrogate signal is sampled densely and overlapping, even at time-points without acquired image information. Because all samples contribute to the loss function, the network learns that there is a continuous relationship between the surrogate signal and internal motion, i.e. the velocities.
Furthermore, provided that the surrogate signals allow to distinguish between inhalation and exhalation, we can model the respiratory motion continuously across all respiratory phases. Therefore, we call our model trajectory-aware.

Commonly used approaches such as normalized cross-correlation (NCC) and bending energy can be chosen as distance measures and regularizers~\citep{wolterink_implicit_2022}. To enforce a plausible, smooth and continuous respiratory motion model, we propose a more complex bio-physically motivated spatial regularizer and an additional temporal regularization that exploits the Euler-based temporal sampling in Eq.~(\ref{eq:integration}).

\subsection{Physics-Enhanced Spatial and Temporal Regularization}\label{subsec:regularizer}
Inspired by PINNs~\citep{raissi_physics-informed_2019}, we aim to tackle the data shortage by utilizing physics-inspired regularization.
We assume the lung \jan{and tumor} tissue to behave like a neohookean hyperelastic material, which closely mimics its natural properties~\citep{concha_micromechanical_2018} and can be formulated as~\citep{arratia_lopez_warppinn_2023}:
\begin{equation}
    \mathcal{R}_{ph}(\varphi^{\bm{\theta}}) = \mathrm{trace}(\mathbf{C}) - 3 - \mathrm{log}(|\mathbf{J}|^2) + \lambda (|\mathbf{J}|-1)^2
    \label{eq:warppinn}
\end{equation}
where $|\mathbf{J}|$ is the determinant of the jacobian $\mathbf{J} = d\phi^{\bm{\theta}}/d\mathbf{x}$, $\mathbf{C}$ is defined as $\mathbf{C} = \mathbf{J}^T\mathbf{J}$, and $\lambda$ is a hyperparamter.
Applying $\mathcal{R}_{ph}$ as regularization in Eq.~\ref{eq:integrated_loss} as proposed in our previous work~(Sec.~\ref{subsec:integratedmodel}) creates a physics-enhanced INR, which is subjected to physically valid displacements in the space domain $\Omega$ at times $t_j$ available in the dataset.
Notably, only utilizing the $\mathcal{R}_{ph}$ regularization does not guarantee any smoothness in the time domain, which neglects the naturally smooth respiratory motion characteristics. Thus, we propose to extend the regularizer in our new trajectory-aware motion model~(Sec.~\ref{subec:diffeo_INR}) by a temporal constraint that guarantees temporal smoothness of the velocity field and physical plausibility of the resulting deformations. Specifically, we propose to add a total variation regularization in the temporal domain to the velocities:
\begin{equation}
    \mathcal{R}_t(v^{\bm{\theta}}) =\frac{1}{2} \int_{t_r}^{t_k} \sqrt{\|\nabla_t v^{\bm{\theta}}\|_2^2 + \epsilon}\ dt
    \label{eq:temp_regul}
\end{equation}
where $\nabla_t$ is the temporal gradient and $\epsilon$ a small constant. Eq.~\ref{eq:temp_regul} allows for sharp edges along time, which is particularly important for the respiratory motion at the end-inhale and end-exhale breathing states. The complete spatio-temporal regularizer is then defined as follows with $\alpha_{t}$ and $\alpha_{ph}$ being hyperparameters:
\begin{equation}\label{eq:regularizer}
    \mathcal{R}_{st}(\varphi^{\bm{\theta}}, v^{\bm{\theta}}) = \alpha_{ph}\mathcal{R}_{ph}(\varphi^{\bm{\theta}})+\alpha_{t}\mathcal{R}_{t}(v^{\bm{\theta}}).
\end{equation}

\section{Experimental Setup}

The purpose of our experimental setup is to investigate the interpolation and extrapolation capabilities of the trajectory-aware motion model. Specifically, the interpolation setup is intended to investigate how well the motion model performs for breathing states within the range of the surrogate signal in the training data, but without including the corresponding image in the training of the model. The extrapolation setup aims to investigate the motion model's performance for breathing states outside the range of the surrogate signal in the training data. The corresponding images are also not included in the training data.

\subsection{Data}
\begin{figure}[b]
    \centering
    \includegraphics[width=\linewidth]{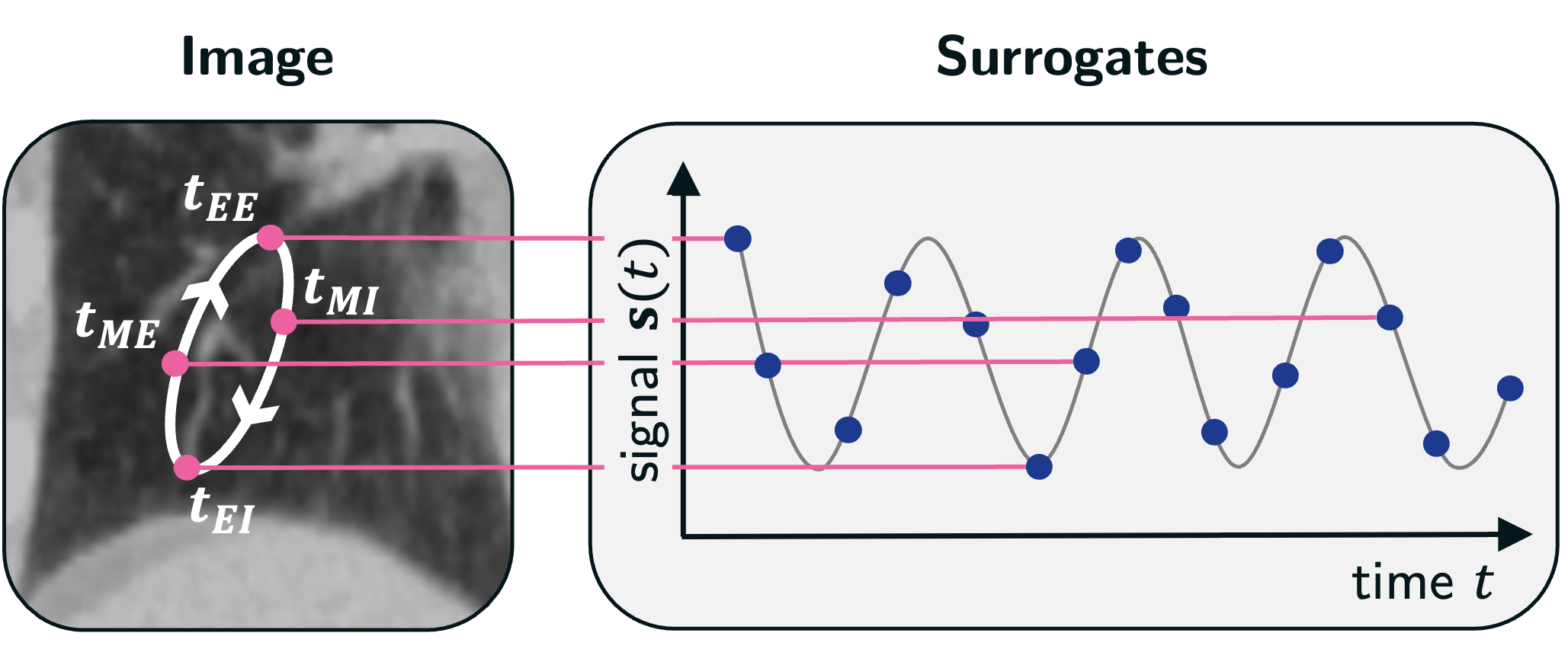}
    \caption{Exemplary motion trajectory of a single voxel tracked over time (left) with a corresponding surrogate signal (right).}
    \label{fig:breathing_cycle}
\end{figure}

For our experiments, we use \jan{a subset of the in-house CT} dataset from \cite{wilms_multivariate_2014}, that contains 4D upper-body CT images of eleven lung cancer patients. For every patient either 10 or 14 images are available at equidistant breathing stages between the inhalation and exhalation state. Specifically, one image is recorded at the end-inhalation (EI) state, one image is recorded at the end-exhalation (EE) state, and the remaining images \jan{are evenly split among the inhalation and exhalation phases and
sorted based on the amplitude}.
\jan{The original dataset contains spirometry measurements of air volume and air flow that were recorded simultaneously with the images (see \cite{yamamoto_reproducibility_2012} and \cite{ehrhardt_optical_2007} for details on image acquisition and image reconstruction). However, to maintain comparability in our experiments, we use simulated spirometry measurements that were calculated retrospectively from the image data and provided by \cite{wilms_multivariate_2014}:}
$\mathbf{s}(t)=\left (V(t), \partial V(t) /\partial t\right)^T$. The lung volumes are extracted and normalized per patient, then bootstrapped by the lung volume at maximum exhalation, which is defined as $V=0$. Furthermore, 70 to 90 corresponding landmarks are manually set per patient at the end-inhale (EI), end-exhale (EE), mid-inhale (MI) and mid-exhale (ME) states (see Fig.~\ref{fig:breathing_cycle} with a corresponding surrogate signal), where the mid breathing states are located at the center between the end-inhale and end-exhale states in the corresponding breathing phases. All images entail visible cancerous tissue.

\subsection{Experiments}\label{subsec:experiments}
Our experimental setup simulates a patient-specific scenario to learn the patient’s breathing pattern during a training phase, enabling the prediction of respiratory motion based on a given surrogate signal during inference.
In order to evaluate the performance of our method for accurate respiratory motion modeling, we design a leave-out-based setup to simulate interpolation and extrapolation scenarios.
Using this particular leave-out setup allows for ground truth evaluation of the predicted respiratory motion, since landmarks are given for the left-out images.
This scenario is like the real-world case of predicting breathing movements outside the known patient-specific range of variability (i.e. extrapolation). By excluding multiple breathing states, we can estimate the model's performance in more extreme cases and assess its robustness. Moreover, utilizing this experiment design allows for a direct performance comparison to the available baseline for the used data~\citep{wilms_multivariate_2014, palm_surrogate-based_2025}.
The extrapolation scenario is particularly important for the clinical application of a respiratory motion model, because respiratory motion is highly variable (e.g. day-to-day, breath-to-breath variability) and the motion model must perform well during inference time on unseen surrogate signals to ensure the radiotherapy effectiveness. 
PRISM-RM is specifically designed to improve performance in this extrapolation scenario, where the spatio-temporal regularization combined with a diffeomorphic setting targets improved smoothness and physical plausibility over time.

\begin{figure}[b]
    \centering
    \begin{tabular}{ccc}
        \includegraphics[width=0.28\linewidth]{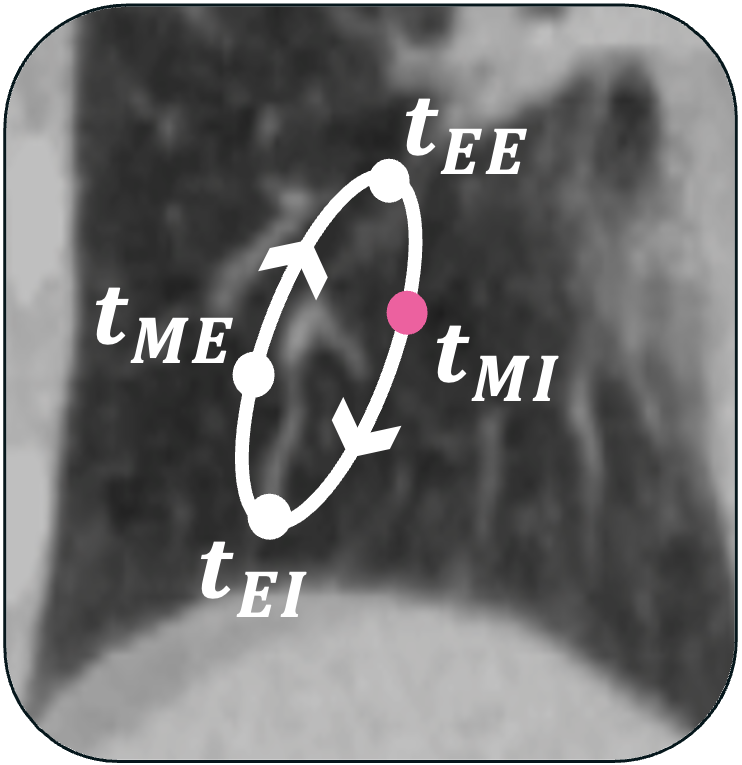} &\includegraphics[width=0.28\linewidth]{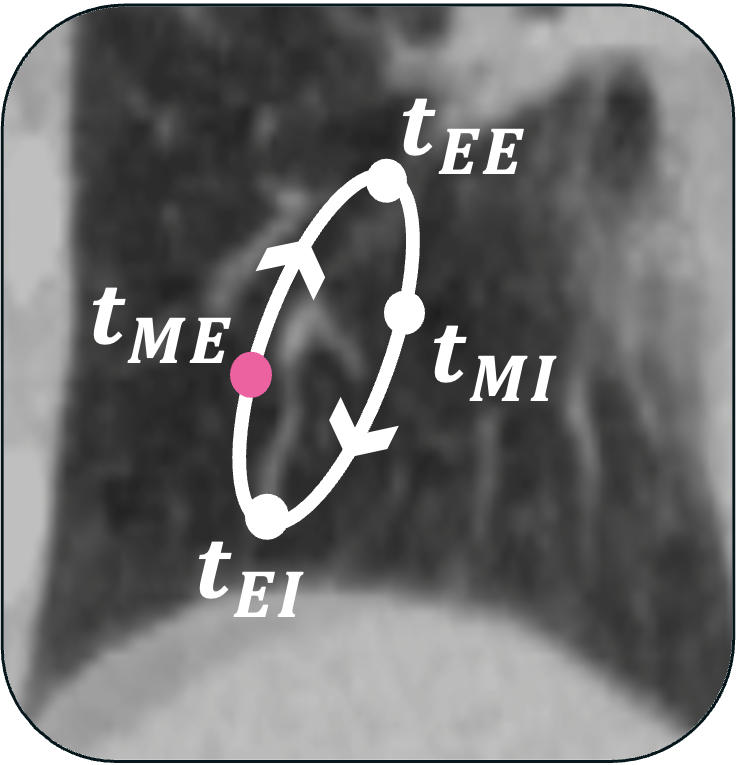} &\includegraphics[width=0.28\linewidth]{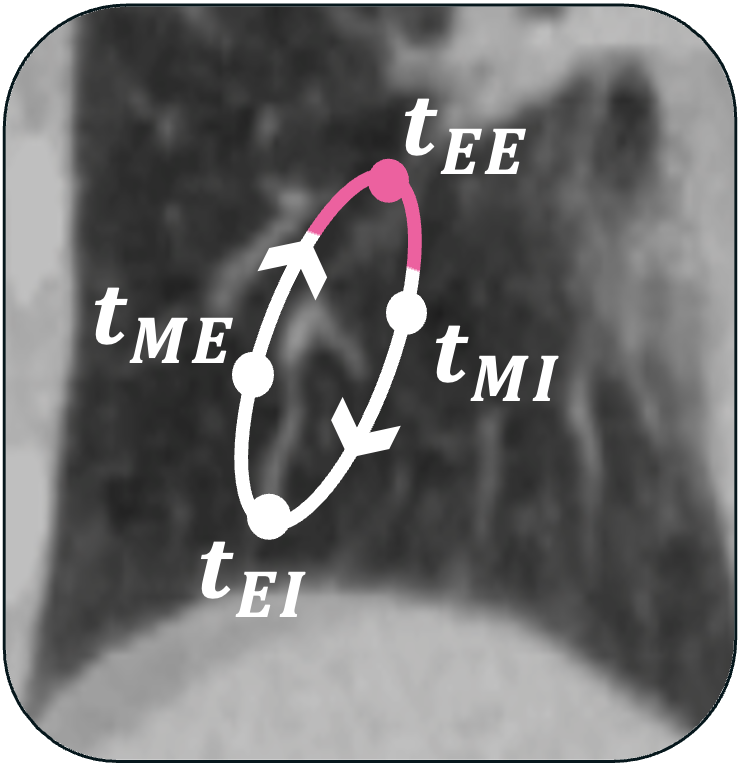} \\
        (a) EI $\rightarrow$ MI & (b) EI $\rightarrow$ ME & (c) EI $\rightarrow$ EE \\
        \jan{Interpolation} & \jan{Interpolation} & \jan{Extrapolation}\\
    \end{tabular}
    \caption{Visualization of the three experiments showing the respiratory states included in the training data in \jan{white} and the excluded one(s) in pink.}\label{fig:experiments}
\end{figure}

Three experiments were performed in which correspondence models were applied to estimate patient-specific motion between end-inspiration and mid-inspiration (EI$\rightarrow$MI), mid-expiration (EI$\rightarrow$ME), and end-expiration (EI$\rightarrow$EE).
Notably, the trajectory-aware approach allows to use any image as a reference for evaluation due to its reference-free nature. However, to ensure comparability, we choose the EI breathing state~\citep{wilms_multivariate_2014, palm_surrogate-based_2025}.  
In all three experiments we use the same dataset per patient but with different training/test data splits. The two interpolation experiments (EI$\rightarrow$MI and EI$\rightarrow$ME) are evaluated by leaving out the phases at mid-expiration (ME) and mid-inspiration (MI) during training. 
The extrapolation experiment (EI$\rightarrow$EE) is designed to evaluate the extrapolation ability of the approaches, i.e. in cases where the respiratory motion lies outside the range of the acquired training images. To make the extrapolation scenario more severe, the neighboring breathing states are omitted during training in addition to the EE breathing state. 
Specifically, if 10 images are available for a patient one additional breathing state in each breathing phase is left out, in case 14 are available two are left out.
A visual explanation of our experiments is represented in Fig.~\ref{fig:experiments} showing the respiratory states included in the training data of the respective experiment in \jan{white} and the excluded one(s) in pink. Here, the trajectory of a single landmark over the entire breathing phase is shown. Where, specifically, for the extrapolation experiment (EI$\rightarrow$EE) the breathing states MI and ME are part of the training set and EE is used for testing.

For performance evaluation in validation and test time, the target registration error (TRE) between the corresponding landmarks is used.
Specifically, for evaluation during testing, the TRE between the landmarks of the warped EI breathing state and the leftout breathing state is used.


As the baseline method we use the sequential two-stage model proposed by~\cite{wilms_multivariate_2014}, which yields state-of-the-art performance for respiratory motion modeling. We then compare our INR-based integrated motion model (Sec.~\ref{subsec:integratedmodel}) and the trajectory-aware approach (Sec.~\ref{subec:diffeo_INR}) to it.

\subsection{Implementation Details} 

\begin{figure}[b]
    \centering
    \includegraphics[width=\linewidth]{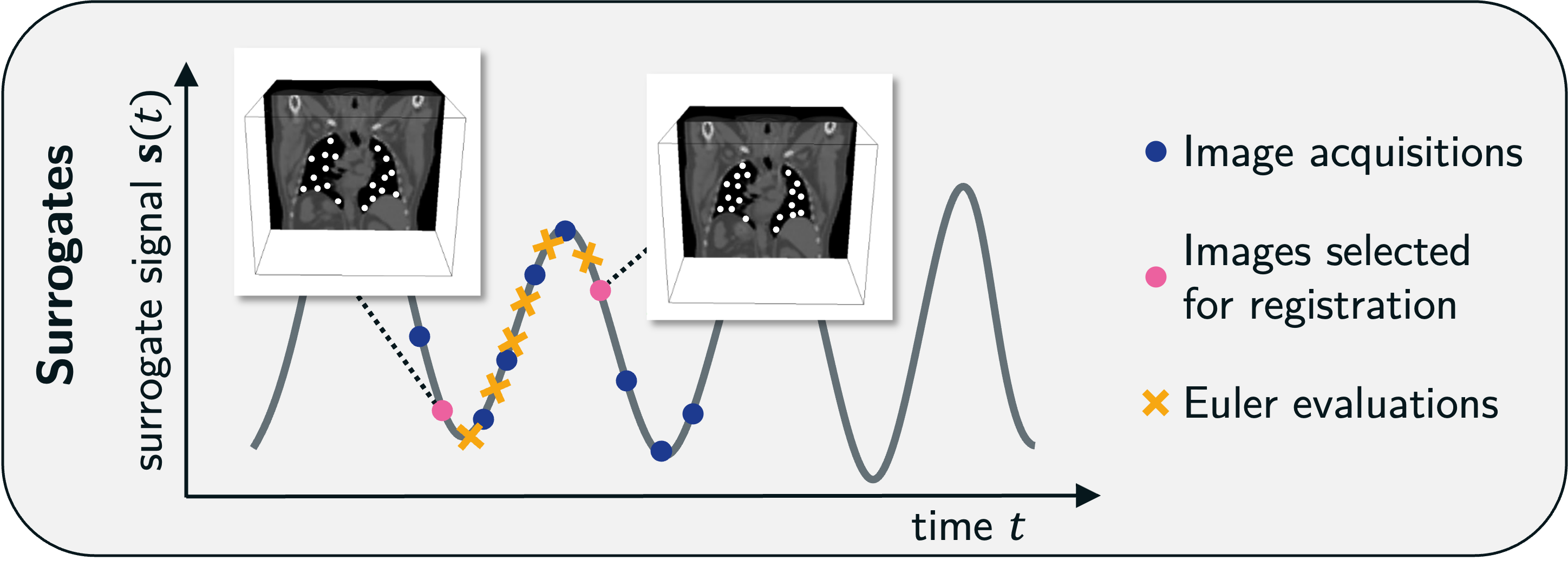}
    \caption{Registration of two images during one optimization step using an Euler scheme. Euler evaluations (yellow dots) can be made in a continuous manner between the actual image acquisitions (blue dots), allowing for the registration of two arbitrary images.}
    \label{fig:euler_steps}
\end{figure}

The velocity fields are represented by a sinusoidal activated INR~\citep{sitzmann_implicit_2020}, which is composed of three hidden layers with 256 neurons each. During training at every epoch we sample 10.000 points in $\Omega \times \mathcal{S}$. Coordinates are sampled using a weighted image-gradient based strategy in the lung region of the image at the end-inhale breathing state of each patient. The surrogate signals of the 10.000 points per epoch are sampled using two different strategies. 40\% are taken from the surrogate signals $\mathcal{S}$ available for images $I_j$ in the dataset (see the blue points in Fig.~\ref{fig:euler_steps} and pink points for a specific registration pair). For the other 60\% the surrogate signal is uniformly sampled over the entire patients breathing cycle (see the yellow points in Fig.~\ref{fig:euler_steps}). Thus, for this 60\% split no image data is available. Specifically, the volumes $V(t)$ sampled uniformly over the entire breathing cycle and the corresponding derivatives are interpolated from the real-world surrogate $s_{\partial V(t)/\partial t}$. The interpolation is done in a piece-wise linear fashion between the known derivatives at the breathing states available in the dataset. We use the normalized cross-corelation $\mathcal{D}(\cdot) = \mathrm{NCC(\cdot)}$ as a similarity measure for the inputs where fixed and moving image are available in dataset. For the 60\% of points at times without available images, no image-based similarity measure is calculated and used for optimization. However, the regularizer is applied to all 10.000 points, because it does not require any image information. During training we validate on the TRE between the EI breathing state and all wrapped images with available landmarks. The regularizer as described above is applied to all points and \jan{its} weighting hyperparameters (Eq.~(\ref{eq:regularizer})) are chosen to $\alpha_{ph}=0.001$ and $\alpha_{t}=0.1$. The $\lambda$ of the physics loss (Eq.~(\ref{eq:warppinn})) is chosen to $\lambda = 1$ as in~\citep{arratia_lopez_warppinn_2023}. The learning rate is set to $1\times10^{-5}$.

\section{Results}

\begin{table*}[t]
\centering
\caption{Results of three registration scenarios measured as TRE in millimeters. \jan{In bold we show the best performance among the two INR-based methods. Presented} methods: `Init' initial alignment before registration; `Seq' registration by a sequential two-stage approach~\citep{wilms_multivariate_2014}; `Integ' registration by our proposed integrated INR approach (Sec.~\ref{subsec:integratedmodel}); `PRISM-RM' registration with the new trajectory-aware approach (Sec.~\ref{subec:diffeo_INR}).}\label{tab:res}

\begin{tabular*}{\textwidth}{@{\extracolsep\fill}c|cc:c@{\hspace{2.5mm}}c@{\hspace{2.75mm}}|cc:c@{\hspace{2.5mm}}c@{\hspace{2.75mm}}|cc:c@{\hspace{2.5mm}}c}
	\hline
    \multirow{2}{*}{ID} & \multicolumn{4}{c}{\textbf{EI} $\rightarrow$ \textbf{MI}} & \multicolumn{4}{c}{\textbf{EI} $\rightarrow$ \textbf{ME}} & \multicolumn{4}{c}{\textbf{EI} $\rightarrow$ \textbf{EE}}\\
    & Init & Seq & \hspace{0.2mm} Integ & PRISM-RM & Init & Seq & \hspace{0.2mm} Integ & PRISM-RM & Init & Seq &\hspace{0.2mm} Integ & PRISM-RM\\ 
    \hline
    01 & 3.98 & 1.68 & \hspace{0.2mm} \textbf{1.96} & 2.15 & 2.08 & 1.48 & \hspace{0.2mm} \textbf{1.54} & 1.86 & 4.25 & 1.29 & \hspace{0.2mm} 2.54 & \textbf{2.10} \\
    02 & 5.08 & 1.65 & \hspace{0.2mm} 2.10 & \textbf{1.92} & 2.51 & 1.32 & \hspace{0.2mm} 1.97 & \textbf{1.82} & 6.29 & 1.56 & \hspace{0.2mm} 2.85 & \textbf{2.17} \\
    03 & 4.43 & 1.83 & \hspace{0.2mm} 1.92 & \textbf{1.72} & 2.41 & 1.83 & \hspace{0.2mm} \textbf{1.99} & 2.68 & 5.45 & 1.74 & \hspace{0.2mm} 2.81 & \textbf{2.41} \\
    04 & 4.95 & 2.24 & \hspace{0.2mm} \textbf{2.08} & 2.27 & 2.52 & 1.77 & \hspace{0.2mm} \textbf{2.14} & 2.34 & 6.19 & 1.76 & \hspace{0.2mm} 3.34 & \textbf{2.13} \\
    05 & 3.76 & 2.42 & \hspace{0.2mm} 2.04 & \textbf{1.70} & 2.41 & 1.78 & \hspace{0.2mm} 2.20 & \textbf{2.02} & 6.79 & 2.34 & \hspace{0.2mm} 4.26 & \textbf{3.98} \\
    06 & 4.56 & 2.15 & \hspace{0.2mm} 1.84 & \textbf{1.73} & 1.98 & 1.87 & \hspace{0.2mm} 1.72 & \textbf{1.65} & 6.44 & 1.46 & \hspace{0.2mm} 4.61 & \textbf{2.51} \\
    07 & 3.80 & 1.78 & \hspace{0.2mm} \textbf{1.74} & 1.81 & 1.92 & 1.97 & \hspace{0.2mm} \textbf{1.82} & 1.96 & 4.31 & 1.61 & \hspace{0.2mm} 2.87 & \textbf{2.51} \\
    08 & 6.68 & 2.59 & \hspace{0.2mm} \textbf{2.11} & 2.22 & 3.84 & 2.08 & \hspace{0.2mm} \textbf{2.16} & 3.46 & 10.76 & 2.66 & \hspace{0.2mm} 5.83 & \textbf{4.38} \\
    09 & 4.11 & 2.02 & \hspace{0.2mm} 1.77 & \textbf{1.60} & 2.76 & 1.70 & \hspace{0.2mm} \textbf{1.77} & 2.86 & 6.40 & 1.74 & \hspace{0.2mm} 3.64 & \textbf{2.85} \\
    10 & 4.03 & 1.68 & \hspace{0.2mm} 1.90 & \textbf{1.81} & 2.07 & 1.43 & \hspace{0.2mm} 1.90 & \textbf{1.62} & 6.06 & 1.60 & \hspace{0.2mm} 4.41 & \textbf{2.63} \\
    11 & 6.71 & 2.24 & \hspace{0.2mm} 2.50 & \textbf{2.19} & 2.94 & 1.77 & \hspace{0.2mm} \textbf{2.00} & 2.50 & 8.31 & 2.65 & \hspace{0.2mm} \textbf{3.07} & 3.45 \\
    \hline
    \hline
    Mean & 4.74 & 2.03 & \hspace{0.2mm} 2.00 & \textbf{1.92} & 2.49 & 1.73 & \hspace{0.2mm} \textbf{1.93} & 2.25 & 6.48 & 1.86 & \hspace{0.2mm} 3.66 & \textbf{2.83} \\
    Std & 1.06 & 0.33 & \hspace{0.2mm} 0.21 & 0.24 & 0.55 & 0.23 & \hspace{0.2mm} 0.20 & 0.58 & 1.81 & 0.47 & \hspace{0.2mm} 1.01 & 0.77 \\
	\hline
\end{tabular*}
\end{table*}

Tab.~\ref{tab:res} presents the results of PRISM-RM, our previously introduced integrated approach and the sequential approach compared for the three leave-out scenarios all on the same dataset.
The integrated INR-based respiratory motion model shows on-par performance with the sequential baseline model for interpolation (ME, MI), but significantly underperforms in extrapolation (EE). PRISM-RM performs comparable to the integrated INR-based respiratory motion model, but outperforms it in extrapolation, \jan{although it still performs worse than the sequential two-stage approach}.
Indeed, a paired T-test reveals no statistically significant differences between PRISM-RM and our previous INR-based integrated approach (Sec.~\ref{subsec:integratedmodel}) in the interpolation scenarios EI$\rightarrow$MI and EI$\rightarrow$ME, but significantly improved results in terms of extrapolation EI$\rightarrow$EE ($p<0.05$).

\begin{figure*}[t]
    \centering
    \begin{tabular}{ccc}
        \includegraphics[width=0.29\textwidth]{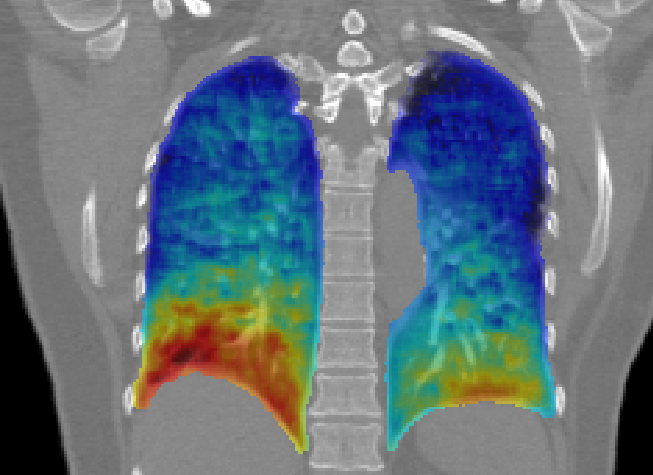} &\includegraphics[width=0.29\textwidth]{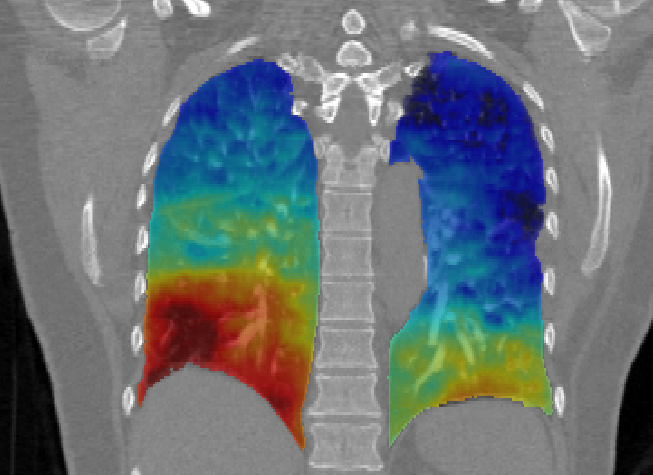} &\includegraphics[width=0.29\textwidth]{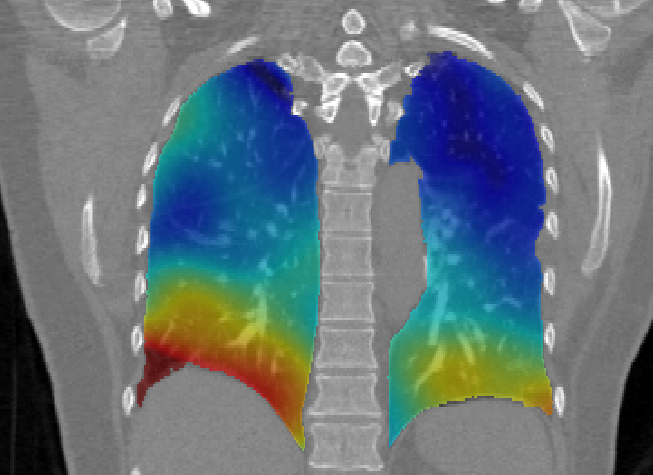} \\
        (a) Intra-patient registration  & (b) Sequential approach & (c) PRISM-RM\\
    \end{tabular}
    \caption{Color-coded magnitudes of the motion (increasing from blue to red) estimated by an image-based intra-patient registration in the ME setting (a); the motion estimation using a sequential surrogate-based registration followed by an ordinary least squares regression~\citep{wilms_multivariate_2014} (b); and the motion estimation by PRISM-RM (c).}\label{fig:res_displacements}
\end{figure*}

Fig.~\ref{fig:res_displacements} shows the magnitude of the resulting displacement fields \jan{for patient six} in the ME experiment, in other words, the displacement magnitude between breathing states EI and ME. \jan{We show patient six as an example, because this patient shows approximately the same TRE improvement between the unregistered images (Init) and the PRISM-RM registered images as the dataset average.} The magnitude increases from blue to red. Note that only intra-patient registration uses ME image information and can therefore be considered a pseudo ground truth. The displacement fields of both models, the sequential and trajectory-aware model, infer the largest displacement at the bottom of the lung. Considering the largest physiological difference between the two breathing states is located at the bottom of the lung (see the red area in \mbox{Fig.~\ref{fig:res_displacements}(a))}, both models infer a plausible displacement field. However, the displacement field magnitude of our trajectory-aware model (\mbox{Fig.~\ref{fig:res_displacements}(c)}) is much smoother than the one of the sequential approach (\mbox{Fig.~\ref{fig:res_displacements}(b)}).

\begin{table*}[t]
    \centering
    \caption{Ablation study results are measured in TRE in millimeters. Compared methods: `Init' before registration; `PRISM-RM' trajectory-aware approach; `S' trajectory-aware approach with only spatial regularization; `T' trajectory-aware approach with only temporal regularization.}\label{tab:ablation}
    \begin{tabular*}{\textwidth}{@{\extracolsep\fill}c|cccc|cccc|cccc}
        \hline
        \multirow{2}{*}{} & \multicolumn{4}{c}{\textbf{EI} $\rightarrow$ \textbf{MI}} & \multicolumn{4}{c}{\textbf{EI} $\rightarrow$ \textbf{ME}} & \multicolumn{4}{c}{\textbf{EI} $\rightarrow$ \textbf{EE}}\\
                & Init & PRISM-RM & S & T & Init & PRISM-RM & S & T & Init & PRISM-RM & S & T\\
         \hline
         Mean   & 4.74 & 1.92 & \jan{\textbf{1.88}} & 1.95 & 2.49 & \jan{\textbf{2.25}} & \jan{\textbf{2.25}} & 2.26 & 6.48 & \jan{\textbf{2.83}} & 2.98 & 2.89 \\
         Std   & 1.06 & 0.24 & 0.23 & 0.24 & 0.55 & 0.58 & 0.44 & 0.42 & 1.81 & 0.77 & 1.01 & 0.9\\
         \hline
    \end{tabular*}
\end{table*}

\begin{table}[t]
    \centering
    \caption{\jan{The average compute time per epoch in seconds during training of the different model types. Compared methods: `PRISM-RM' trajectory-aware approach; `S' trajectory-aware approach with only spatial regularization; `T' trajectory-aware approach with only temporal regularization.}}
    \label{tab:durations}
    \jan{
    \begin{tabular}{c|ccc}
    \hline
            & PRISM-RM & S & T \\
         \hline
         MI & 40.1s & 27.8s & \textbf{21.2}s \\
         ME & 40.3s & 29.7s & \textbf{25.7}s \\
         EE & 18.9s & 13.6s & \textbf{11.2}s \\
    \hline
    \end{tabular}
    }
\end{table}

\subsection{Ablation Study}
We conduct an ablation study to investigate the impact of the two different regularization types (spatial and temporal) on the interpolation and extrapolation performance of PRISM-RM.
Tab.~\ref{tab:ablation} shows an ablation study on the different regularization types of the trajectory-aware model. 
We compare the overall mean and standard deviation of the TRE resulting from registration using both the spatial and temporal regularizers simultaneously (PRISM-RM) or each regularizer independently. 
\jan{In the interpolation experiment EI~$\rightarrow$~MI only using the spatial regularizer performs best by a small margin. For the EI~$\rightarrow$~ME interpolation experiment PRISM-RM performs comparable to our previously introduced only spatially regularized integrated INR approach and also compared to only using the temporal regularizer. Thus, the integration of both a temporal and spatial regularization overall does not necessarily improve the model performance significantly in interpolation scenarios.}
However, for the extrapolation experiment (EI~$\rightarrow$~EE) Tab.~\ref{tab:ablation} shows that the integration of regularization both in space and time improves the performance significantly. Also, the decreasing standard deviation in the extrapolation suggests that the combination also increases PRISM-RM's certainty about the motion. Only using spatial regularization yields worse results compared to solely using temporal regularization in the extrapolation case. This follows the intuition, since the spatial regularization only induces constraints at the individual time steps and not over time\jan{, which is the additional constraint that is necessary for the extrapolation scenario, since the model cannot derive information from the data in this case.}

\jan{The models were trained on a NVIDIA A30. The training times for training and inference of the different experiments in the ablation study are presented in Tab.~\ref{tab:durations}. The epoch duration in the extrapolation experiment is the shortest as the number of breathing states in the dataset is smaller compared to the interpolation experiments (Sec.~\ref{subsec:experiments}). Further, the trajectory-aware approach with only the temporal regularization is faster then the other methods, because fewer derivatives must be calculated due to the missing physics-inspired regularization. Hence, the inference duration for a single image is similar in all experiments, as no additional backpropagation is required for the regularizer. The inference of a model takes around 2.2s for a single image.}

\section{Discussion and Conclusion}

This work explores the potential of INRs in integrated surrogate-based respiratory motion modeling. In our initial work~\citep{palm_surrogate-based_2025} INRs were used to enable the learning of complex spatial and temporal relationships between surrogate and internal motion in an integrated model. This allows for advancing beyond sequential two-stage models and the assumptions of common linear or polynomial correspondence models. The evaluation results show this INR-based model to be equivalent to established two-stage models for interpolation scenarios, but its ability to generalize to unseen breathing states during extrapolation remains limited and requires further methodological improvements.

As a continuation of this work, we present PRISM-RM, a trajectory-aware INR-based framework for integrated respiratory motion modeling that addresses key limitations of our previous work, especially in extrapolation to unseen breathing states. Unlike previous methods constrained by fixed reference states, our proposed PRISM-RM framework represents a time-continuous representation of the entire respiratory cycle and thus allows motion estimation between arbitrary respiratory phases, significantly extending the training basis by using arbitrary image pairs. This is further enhanced by the applied Euler integration along continuous surrogate signals, whereby a consistent and smooth motion is learned even outside of available image information. \jan{Since the lung tissue is modeled as a neohookean hyperelastic material, the smooth deformation is desirable and also corresponds to its natural behavior.} By leveraging spatio-temporal regularization within the diffeomorphic framework, our method enables smooth, physiologically consistent motion fields while maintaining adaptability to patient-specific variations.

Our results show, that PRISM-RM maintains a comparable performance in interpolation tasks while significantly improving extrapolation accuracy compared to the integrated INR-based respiratory motion model, which is critical for unseen breathing patterns during treatment delivery. Our ablation study finds the integration of spatio-temporal regularization, in particular the temporal component, crucial to stabilize predictions across the breathing cycle. This greatly aligns with the physiological continuity of respiratory motion. Notably, the decreasing variance in the extrapolation suggests increased confidence in the trajectory-aware model’s predictions, despite the data limitations inherent to surrogate-based approaches.

Despite the advances of PRISM-RM in INR-based modeling, limitations remain. Although the performance in extrapolation improved, it still lags behind sequential approaches. This indicates that INRs are not yet fully overcoming the well-known generalization barriers \jan{especially in the extrapolation experiment} for unseen respiratory states. 
Thus, further methodological improvements are needed, e.g. we plan to apply neo-Hookean regularization also along the Euler steps, similar to the temporal regularization, to improve the spatial regularization at times where no images are available. Another line of future work will be the exploration of global motion atlases for initialization which afterwards can be conditioned to individual patients and we plan to investigate how the scaling-and-squaring algorithm can be applied to a sparse velocity field to take advantage of its computational effectiveness. Overall, PRISM-RM reveals a promising line of research and is a meaningful step towards clinically applicable and patient-specific respiratory motion modeling, with the potential to improve precision in radiotherapy delivery.



\acks{This work was funded by grant 16IS24006C from the German Federal Ministry of Research, Technology and Space (BMFTR) and by the state of Schleswig-Holstein, Germany, through grant 22024003.}

%
\ethics{The work follows appropriate ethical standards in conducting research and writing the manuscript, following all applicable laws and regulations regarding treatment of animals or human subjects.}

\coi{We declare we don't have conflicts of interest.}

\data{The used dataset was introduced by \cite{wilms_multivariate_2014} and kindly made available for this work. For details on the data availability for third-parties the interested reader is kindly referred to the corresponding author.}

\bibliography{references}

\end{document}